%% file: acl_latex.tex
\pdfoutput=1

\documentclass[11pt]{article}

\usepackage[dvipsnames]{xcolor}
\usepackage[final]{acl}

\usepackage{times}
\usepackage{latexsym}

\usepackage[T1]{fontenc}

\usepackage[utf8]{inputenc}

\usepackage{microtype}

\usepackage{inconsolata}

\usepackage{graphicx}
\usepackage{algorithm}
\usepackage{algpseudocode}

\usepackage{booktabs} 
\usepackage{amsmath, amsthm, amssymb}
\usepackage{bbm}

\usepackage{parskip}
\usepackage{soul}
\usepackage{cleveref}

\usepackage{textcomp}

\usepackage{setspace}
\usepackage{makecell}
\usepackage{xcolor,colortbl}
\usepackage{soul}
\usepackage{color}
\usepackage{CJKutf8}
\usepackage{tcolorbox}\tcbuselibrary{skins}
\usepackage{tipa}
\usepackage{xcolor}

\newcommand{\fon}[1]{\fontfamily{#1}\selectfont} 
\definecolor{CB_pink}{HTML}{FFAABB}
\tcbset{prompt/.style={
    enhanced,
    size=fbox,
    boxrule=3pt,
    arc=2mm,
    auto outer arc,
    left=10pt,
    right=10pt,
    top=10pt,
    bottom=10pt,
    colback=CB_pink!15,
    colframe=CB_pink!30,
    coltitle=CB_pink!25!black, 
    fontupper=\fon{cmtt}, 
}}

\crefformat{appendix}{\S#2#1#3}
\crefformat{section}{\S#2#1#3} 
\crefformat{subsection}{\S#2#1#3}
\crefformat{subsubsection}{\S#2#1#3}

\newcommand{\hlc}[2][yellow]{{%
    \colorlet{foo}{#1}%
    \sethlcolor{foo}\hl{#2}}%
}

\newcommand\yellowcell{\cellcolor[rgb]{1,0.843,0}}
\newcommand{\datasetname}[1]{LessonLink}

\title{Problem-Oriented Segmentation and Retrieval: \\ Case Study on Tutoring Conversations}

\author{Rose E. Wang  \\
  \texttt{rewang@cs.stanford.edu} \\\And
  Pawan Wirawarn \\
  \texttt{pawanw@stanford.edu} \\\AND
  Kenny Lam \\
  \texttt{knlam@stanford.edu} \\\And  
  Omar Khattab \\
  \texttt{okhattab@stanford.edu} \\\And  
  Dorottya Demszky \\ 
  \texttt{ddemszky@stanford.edu} \\ \AND
  Stanford University
  }

\begin{document}
\maketitle
\begin{abstract}
Many open-ended conversations (e.g., tutoring lessons or business meetings) revolve around pre-defined reference materials, like worksheets or meeting bullets.
To provide a framework for studying such conversation structure, we introduce \textbf{Problem-Oriented Segmentation \& Retrieval} (POSR)\footnote{Pronounced as ``poser'' (\textipa{/\textprimstress poz\textschwa r/}), a perplexing problem.}, the task of \textit{jointly} breaking down conversations into segments and linking each segment to the relevant reference item.  
As a case study, we apply POSR to education where effectively structuring lessons around problems is critical yet difficult.
We present \textbf{\datasetname{}}, the first dataset of real-world tutoring lessons, featuring 3,500 segments, spanning 24,300 minutes of instruction and linked to 116 SAT\textregistered{} math problems.
We define and evaluate several joint and independent approaches for POSR, including segmentation (e.g., TextTiling), retrieval (e.g., ColBERT), and large language models (LLMs) methods.
Our results highlight that modeling POSR as one joint task is essential: POSR methods outperform independent segmentation and retrieval pipelines by up to +$76\%$ on joint metrics and surpass traditional segmentation methods by up to +$78$\% on segmentation metrics.
We demonstrate POSR's practical impact on downstream education applications, deriving new insights on the language and time use in real-world lesson structures.\footnote{Our code and dataset are open-sourced at \url{https://github.com/rosewang2008/posr}.}
\end{abstract}

\section{Introduction}
Across education, business, and science, many open-ended conversations like meetings or tutoring sessions are designed to address a set of pre-defined topics. 
As a prominent example, educators often shape their lessons around worksheet problems.
Structuring lessons effectively is critical but challenging, as educators must allocate the right amount of time to different problems, while addressing different student learning needs \citep{haynes2010complete, henderson1997transformative, panasuk2005effectiveness}. 
However, many novices or educators teaching large groups of students struggle with lesson structuring and often run out of time \citep{stradling1993differentiation, pozas2020teachers, deunk2018effective, takaoglu2017challenges, hejji2019study}. 

\begin{figure}[t]
    \centering
    \includegraphics[width=\linewidth]{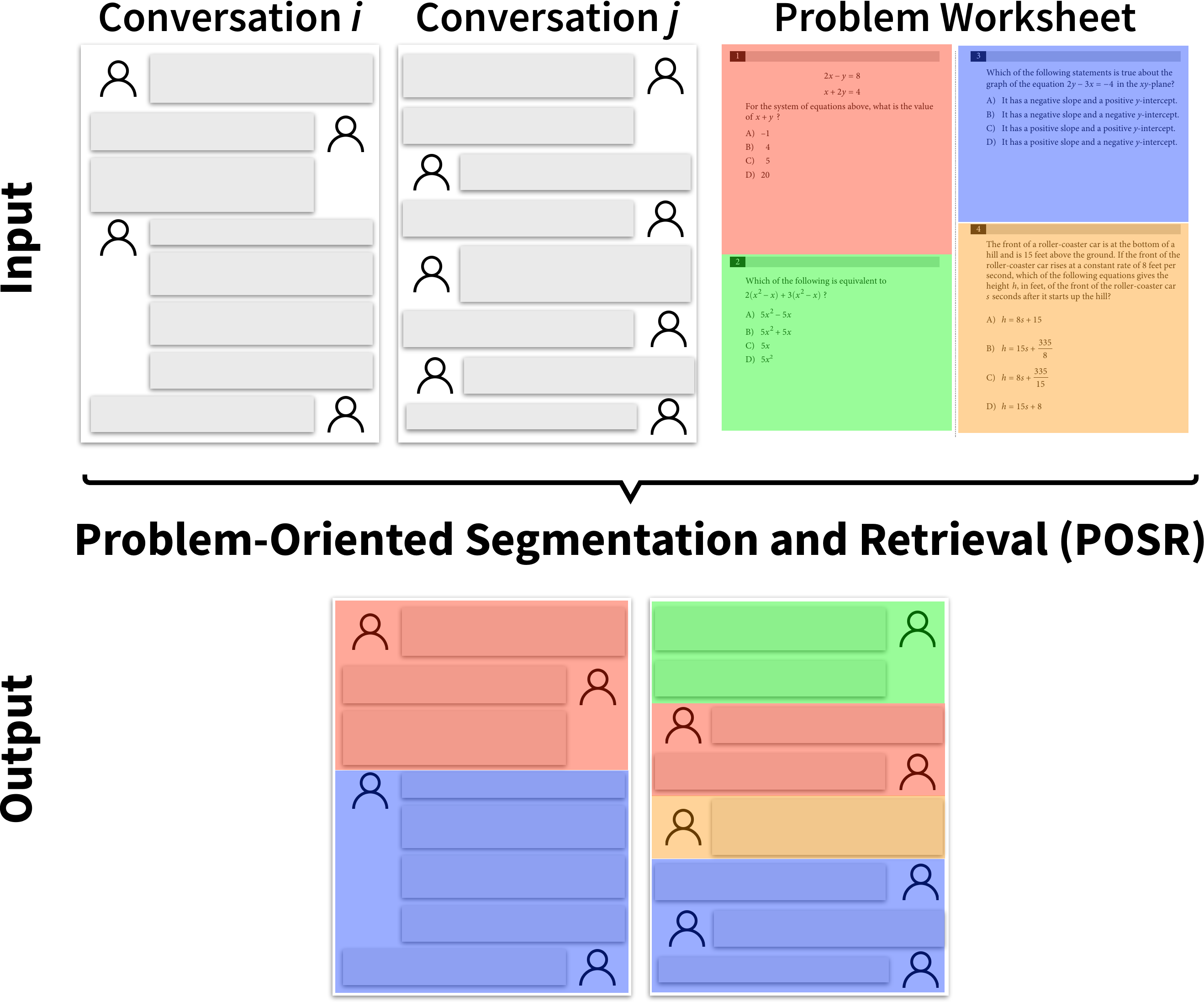}
    \caption{\small{
    Problem-Oriented Segmentation and Retrieval (POSR) provides a framework for studying conversation structure around reference materials.
    For example, while conversations $i,j$ discuss the same worksheet, POSR reveals that conversation $i$ covers fewer problems than $j$ but spends more time per problem.
    }
    }
    \label{fig:example}
    \vspace{-1em}
\end{figure}

\textit{Providing evidence-based insights on lesson structuring} is a key step towards addressing this challenge. 
These insights provide educators feedback on their teaching~\citep{fishman2003linking, kraft2018effect, lomos2011professional, desimone2009improving}, 
tutoring platforms on training priorities~\citep{hilliger2020design, gottipati2018competency, hilliger2022lessons} and curriculum developers on material design~\citep{o2008defining, fullan1977research}. 
Unfortunately, obtaining insights on lesson structures at scale is challenging.

The study of conversation structure around reference materials draws on concepts from two, typically distinct natural language processing (NLP) tasks: \textit{discourse segmentation} to identify segments in the conversations and \textit{information retrieval} (IR) to retrieve the relevant reference material for each segment.
While each task has rich literature, studying them jointly reveals real-world challenges that existing works bypass.
For example, discourse segmentation methods assume that conversations share the same structure~\citep{ritter2010unsupervised, hearst1993subtopic, chen2020multi}, but education conversations have unique structures as teachers adapt their lessons to different needs. 
While prior IR work has studied supporting natural-language queries over conversations \citep{sanderson2010test, oard2004building, chelba2008retrieval}, the reverse task of using open-ended conversation segments as queries for retrieving domain-specific reference materials has not received similar attention.

To address these gaps, we make several key contributions. 
We define the \textbf{Problem-Oriented Segmentation and Retrieval} (POSR) task for jointly segmenting conversations and linking segments to relevant reference materials, such as worksheet problems (Figure~\ref{fig:example}).
Unlike segmentation or retrieval alone, the joint POSR task reflects the realistic opportunities and challenges presented by knowing the potential reference topics (from the reference materials) for conversation segments. 

POSR provides a general framework for studying conversation structure around reference materials. 
As a case study, we apply POSR to the education setting. 
We contribute \textbf{\datasetname{}, a novel dataset of real-world tutoring lessons featuring 3,500 segments, 116 SAT\textregistered{} math problems, and over 24,300 minutes of instruction}. 
Our open-source dataset consists of real tutoring conversations paired with SAT\textregistered{} math worksheets, each conversation lasting about 1.5 hr long.
Each conversation is segmented and each segment is linked with one of the 116 problems. 
To the best of our knowledge, this is the first dataset to include real-world conversations of unique structures linked with reference materials like worksheets.

Evaluating POSR is challenging: Existing segmentation metrics do not measure time-weighed errors and existing metrics fail to reflect the subtle ways in which segmentation and retrieval errors interact.
To address this, we contribute \textbf{time-aware segmentation metrics} adapted from standard line-based metrics (e.g., WindowDiff from \citet{pevzner-hearst-2002-critique}) and introduce the \textbf{Segmentation and Retrieval Score (SRS)} to jointly measure segmentation and retrieval accuracy as the proportion of conversation where the retrieved item matches the ground truth.

We \textbf{define and evaluate a suite of segmentation, retrieval and POSR methods} on \datasetname{}, including traditional segmentation methods like TextTiling \citep{hearst1997text}, popular IR methods like ColBERT \citep{khattab2020colbert} and  long-context large language models (LLMs) like Claude and GPT-4 \cite{anthropic2024claude,openai2024gpt}.
Our results highlight the importance of POSR's joint approach:
POSR methods outperform independent segmentation and retrieval pipelines by up to +$76\%$ on SRS metrics and traditional segmentation methods by up to +$78$\% on segmentation metrics. 
However, several challenges remain. In domains with high privacy risks like education, companies are often unwilling to share data long-term due to privacy concerns. 
Moreover, while LLMs achieve strong POSR performance, their high API costs on long texts raise scalability concerns. 
Our findings motivate the need for more cost-effective, open-sourced methods that can deliver high accuracy on joint reasoning tasks like POSR.

Finally, to further highlight the utility of POSR to real-world scenarios, we describe \textbf{two novel applications of POSR} to illustrate its potential for impacting evidence-based practices in education.
First, through a linguistic analysis, we discover that tutors who spend more time on problems provide richer conceptual explanations.
Tutors who spend less time provide procedural explanations. 
Second, POSR quantifies wide variability in how long tutors spend on the same problem.
These examples point to opportunities for improving language and time-management practices.

\section{Related Work}
\textbf{Discourse segmentation} is the task of partitioning conversations into segments, traditionally a pre-processing step before retrieval or summarization of conversations \citep{hearst1993subtopic, callan1994passage,wilkinson1994effective,galley2003discourse, chen2020multi, althoff2016large, salton1991automatic,salton1991global,salton1996automatic, huang2003applying}. 
Different domains like customer service or meetings define segments differently, e.g. as a speech act, a topic, or a conversation stage \citep{liu-etal-2023-joint, riedl2012topictiling,prabhakaran2018detecting}; 
In this work, we study \textit{problem-oriented} segments: conversation segments that discuss individual math problems. 
While most existing segmentation methods assume conversations exhibit predictable structure \citep{ritter2010unsupervised, hearst1993subtopic, chen2020multi}, education conversations are diverse and lack such predictable structure.

\textbf{Math information retrieval} poses special challenges  \citep{munavalli2006mathfind, sojka2011art, nguyen2012math} because math expressions can be difficult to represent contextually \citep{schubotz2016semantification, kamali2013retrieving, zanibbi2012recognition, aizawa2021mathematical}. Our setting combines these challenges with the additional difficulty of treating conversational segments as queries, unlike typical retrieval using well-formed keyword queries \citep{wang-etal-2024-backtracing}. 
Our \datasetname{} dataset provides a new resource of real-world education conversations segmented and linked to math problems from worksheets.
This enables the study of POSR, combining discourse segmentation with retrieval of math materials.

\textbf{Evaluation metrics} for segmentation include $P_k$~\cite{beeferman1997text} and WindowDiff~\cite{pevzner-hearst-2002-critique}.
Both measure the segmentation accuracy based on a \textit{line-level} sliding window \citep{morris1991lexical,kozima1996text,reynar1999statistical,choi2000advances,beeferman1999statistical} but neither account for the time duration of a line, which can confound accuracy reporting for real-world applications  \citep{grosz1992some,nakatani1995discourse,passonneau1997discourse,hirschberg1998acoustic, repp2007segmentation}. 
We develop a time-based version of $P_k$ and WindowDiff and propose a time-based SRS metric for assessing the holistic performance.

\section{Problem-Oriented Segmentation and Retrieval (POSR) \label{sec:task}}
We define the task of Problem-Oriented Segmentation and Retrieval (POSR) as jointly dividing a \textit{conversation transcript} into segments and retrieving the \textit{relevant topic} (e.g., problem) discussed in each segment. 
While segmentation and retrieval are individually challenging, POSR jointly addresses them together to improve ecological validity and expose new system design tradeoffs.
We hypothesize (and show in Section~\cref{sec:results}) that systems aware of retrieval topics will segment better, and vice versa, motivating joint POSR methods. 

\subsection{Task Definition}
Given a transcript $T = \langle T_1, ..., T_N \rangle$ of $N$ lines and a corresponding reference corpus $R = \langle R_1, ..., R_W\rangle$ (e.g., a worksheet of problem entries), the POSR objective is to output an array of segment id and problem reference id for each line in the transcript, $Y = [(s_1, w_1), (s_2, w_2), \dots, (s_N, w_N)]$: 

\begin{itemize}
    \item $s_1, \dots, s_N$ is the segment id for each line in line. So, $s_1$ is the segment id for the line 1, $s_2$ the segment id for line 2, and so on. 
    \item $w_1, \dots, w_N\in \{R_1, \dots, R_W\} $ indicate the problem reference id from the corpus.\footnote{If $s_i = s_j$ then $w_i=w_j$.}
\end{itemize}

Since these transcripts originate from real-world conversations, each line $T_i$ is associated with a start and end timestamp, $t_i^{\text{start}}, t_i^{\text{end}}$.
Algorithm~\ref{alg:posr} highlights \colorbox{blue!30}{POSR methods}, which take both transcript $T$ and retrieval corpus $R$ into account for segmentation, in contrast to \colorbox{red!30}{independent} segmentation and retrieval methods.

\begin{algorithm}[t]
\begin{algorithmic}
\small
\Require $T, R$
\If{ with POSR }
    \State \colorbox{blue!30}{$s_1, \dots, s_N \gets \texttt{segment}(T, R)$}
\Else{} 
    \State \colorbox{red!30}{$s_1, \dots, s_N \gets \texttt{segment}(T)$}
\EndIf
\State $w_1, \dots, w_N \gets \texttt{retrieve}([s_1, \dots, s_N], R)$
\end{algorithmic}
\caption{POSR vs. non-POSR methods}\label{alg:posr}
\end{algorithm}

\subsection{Metrics}
To evaluate the effectiveness of POSR methods, we introduce the standard and our novel metrics for evaluating segmentation and retrieval individually and jointly.
As evident in Algorithm~\ref{alg:posr}, the segmentation metrics help capture how segmentation may be improved by accounting for the retrieval corpus. 
We additionally adapt standard metrics to also take time into account.
Finally, we also account for practical considerations by reporting cost. 

\paragraph{Existing, line-based segmentation metrics.} 
We use two established metrics for segmentation accuracy: WindowDiff from \citet{pevzner-hearst-2002-critique} and $P_k$ metric from \citet{beeferman1999statistical}. 
Both use a line-based sliding window approach that measures boundary mismatches within the window. 
Lower values are better for both metrics. 
For example, WindowDiff is computed as:
\begin{align*}
    &\text{WindowDiff}(Y, Y^*) =  \\
    &\frac{1}{N-k} \sum^{N-k}_{j=1} \mathbbm{1}(|b(s_{j:j+k}) - b(s^*_{j:j+k})| > 0), 
\end{align*}
\vspace{-1em}

where $b(\cdot)$ represents the number of boundaries within the $\cdot$ window and $k$ is typically set to half of the average of the true segment line size.
$P_k$ is similar but penalizes false-negatives more, i.e., missed segments.
For conciseness, we leave $P_k$'s definition in Appendix \cref{sec:pk}.

\paragraph{New, time-based variants of segmentation metrics.} 
Existing segmentation metrics operate at a line-level and do not account for the time duration of segments. 
However, in education settings, time spent per segment is crucial to understanding lesson structures \citep{stevens2012role, martens2010attentional, heim2012developmental,eze2017lecture}.
To address this, we propose Time-WindowDiff and Time-$P_k$, new \textit{time-based} variants of $P_k$ and and WindowDiff.
Time-Windowdiff is calculated as:
\begin{align*}
    \text{Time-WindowDiff}&(Y, Y^*) =\\
    \frac{1}{N-k}\sum^{N-k}_{j=1}  \mathbbm{1}(|b(&s_{t_j^{\text{start}}:t_j^{\text{end}}+\Delta_k})  \\ 
    & - b(s^*_{t_j^{\text{start}}:t_j^{\text{end}}+\Delta_k})| > 0), 
\end{align*}

where $\Delta_k$, the time duration of the sliding window, is half of the average true segment duration (similar to $k$).
$b(s_{t_j^{\text{start}}:t_j^{\text{end}}+\Delta_k})$ refers to the number of boundaries within the window that starts at $t_j^{\text{start}}$ and ends at $t_j^{\text{end}}+\Delta_k$.
This ensures that long and short segment durations are appropriately weighted in the evaluation.
For conciseness, we leave Time-$P_k$'s definition in Appendix \cref{sec:pk}.

\paragraph{API cost.} 
Closed-sourced models result in high API usage costs, especially on thousands of long conversations such as in our setting.\footnote{Third-party models additionally raise privacy and intellectual property concerns especially in domains that deal with sensitive data, like student data and copyrighted materials.}
Educational organizations may be less inclined to rely on expensive methods without justified trade-offs. 
Thus, we report the average cost per 100 transcripts\footnote{Based on OpenAI and Anthropic pricing in 05/24-06/24.}.

\paragraph{The Segmentation Retrieval Score (SRS).} 
Evaluating POSR methods presents unique challenges because of interdependencies between segmentation and retrieval. 
On the one hand, segmentation may improve with access to the retrieval corpus in disambiguating segment boundaries.
On the other hand, incorrect segmentation make retrieval evaluations difficult as the retrieved content cannot be easily checked with misaligned segment boundaries and IDs. 

We propose the Segmentation Retrieval Score (SRS), which accounts for this by evaluating the correctness of retrieved topics, conditioned on the predicted segmentation. 
False positive segments overly penalize an exact segment match.
Therefore, SRS only requires the retrieved topic $w_j$, determined based on the predicted segment $s_j$ (rf. Algorithm~\ref{alg:posr}), to match the reference $w_j^*$ for a line to be considered correct. 
This allows some flexibility in segment boundaries as long as the retrieved topics are accurate.
SRS is defined as: 

\vspace{-2em}
\begin{align*}
    \alpha\text{-SRS}(Y, Y^*) = \frac{1}{\sum_j \alpha_j }\sum_{j=1}^{N} \alpha_j \mathbbm{1}({w_j(s_j) == w^*_j})
\end{align*}
\vspace{-1em}

where line-based SRS has $\alpha_j = 1$ and time-based SRS has $\alpha_j = t^{\text{end}}_{j} - t^{\text{start}}_{j}$.   

\section{The \datasetname{} Dataset \label{sec:dataset}}
We introduce the \datasetname{} dataset as a concrete case study of POSR. 
\datasetname{} contains real-world tutoring lesson transcripts segmented and linked with problems in SAT\textregistered{} math worksheets. 
The dataset features 3,500 segments of over 24,300 minutes of instruction, featuring 1,300 unique speakers and 116 linked problems.
Table~\ref{tab:dataset_statistics} summarizes the statistics of the dataset.
We release the \datasetname{} dataset under the CC Noncommercial 4.0 license\footnote{\url{https://creativecommons.org/licenses/by-nc/4.0/}}.

\input{figures/dataset_statistics}

\paragraph{Data source.} We collected the data in partnership with \href{https://schoolhouse.world/}{Schoolhouse.world}, a free peer-to-peer tutoring platform that supports over $\sim$80k students worldwide with the help of $\sim$10k volunteer tutors. 
One of their main focuses is to help high school students prepare for the SAT, a standardized test used for college admissions in the United States. 
The platform shared de-identified transcripts with us from their March 2023 SAT\textregistered{} Math Bootcamp, a four week-long course where tutors met with students in small groups twice a week to practice SAT\textregistered{} math problems. 
We randomly picked 300 transcripts.
Schoolhouse received consent from parents and students to share de-identified data for research purposes. 
The maximum tutor-student ratio in each bootcamp is 1:10. 
Tutoring lessons are 80 minutes long.
Schoolhouse recommends a lesson structure that starts with 30 minutes of warm-up exercises followed by the students working on the worksheet independently and then a group review.
Tutors have freedom in structuring their lesson and they typically use their students' practice test results to determine what to focus on. 

\paragraph{Transcripts.} 
Each tutoring lesson is recorded and transcribed automatically via Zoom. 
Schoolhouse de-identified the transcripts using the Edu-ConvoKit library~\citep{wang2024convokit}, with tutor and student names replaced with placeholder tokens ``[TUTOR]'' and ``[STUDENT]''. 

\paragraph{Worksheets.} 
Each transcript is linked to an SAT\textregistered{} problem worksheet that the tutor and students work on during the lesson. 
The sheets include official, publicly available math practice problems created by College Board\textregistered, the organization that administers the SAT\textregistered{} exams.\footnote{\url{https://satsuite.collegeboard.org/sat/practice-preparation/practice-tests}}
Each worksheet has about 16 problems on average.
We split each worksheet into separate problem images, and use Pytesseract, an optical character recognition (OCR) tool, to extract the text content from the images \citep{pytesseract2017}. 
OCR does not capture the visual components (e.g., graphs). 
We focus only on using the text data, and leave visual data for future work. 

\paragraph{Annotation.} 
The definition of a segment varies across domains like customer service, meetings, and tutoring sessions \citep{liu-etal-2023-joint, riedl2012topictiling}.
Our definition builds on Schoolhouse.world's curriculum structure that dedicates time for an introduction to the session, targeted warm-up exercises, and worksheet problems. 
We use the following segment categories:
(1) \textbf{Informal.} These segments include introductory talk or off-task discussions
\citep{carpenter2020detecting,rodrigo2013student}. Examples include the group doing an ice-breaker game.
(2) \textbf{Warm-up problem.} These segments discuss warm-up problems that are not a part of the session's main worksheet. 
(3) \textbf{Worksheet problem.} These segments discuss a problem from the session's main worksheet.

We recruited 3 annotators who were familiar with the Schoolhouse materials and tutoring session structure. 
This domain familiarity was important in ensuring high-quality annotations. 
The annotation process was carried out using Excel sheets, and annotators were compensated at a rate of \$20 per hour.
Segment annotations happen at the level of a transcript line, as provided by Zoom. 
Each transcript line includes a start and end timestamp in milliseconds. 
While Zoom uses its own proprietary ASR technology, the lines typically capture a single utterance without the speaker making a pause. 
To ensure alignment and consistency, the start/end of a segment happens on the end of a sentence. 
This means that if a sentence is broken up into two lines, the last line would be considered for the segment annotation.

To determine human agreement on this task, the annotators annotated the same 30 lesson transcripts for segments and linked problems. 
On a line-level, the inter-rater segmentation accuracy was 98.9\% and retrieval accuracy was 100\%. 
We also use Cochran's Q \citep{cochran1950comparison} to evaluate segmentation agreement, similar to prior work \citep{galley2003discourse}: Cochran's test evaluates the null hypothesis that the number of subjects assigning a boundary at any position is random. 
The test shows that the inter-rater reliability is significant to the $0.01$ level for $98\%$ of the transcripts.
Given the high inter-rater agreement, the 3 annotators annotated 300 transcripts. 
We create a small 1:10 train/test split on our dataset: The train set containing 30 transcripts and the test set 270 transcripts. 
We intentionally have a large test set: 
While some methods require a training set, we prioritize a robust  evaluation of zero-shot methods and thus have a larger test set.
This approach is consistent with other zero-shot evaluations in the literature \citep{chen2021evaluating, wang-etal-2024-backtracing}, where large test sets are used for robust comparison of zero-shot methods.

\section{Evaluation \label{sec:evaluations}}
This section describes the methods and evaluation setup which uses \datasetname{}'s test split. 
Appendix~\cref{sec:prompts} includes more information on our prompting setup for GPT4 and Claude LLMs.

\paragraph{Segmentation.} We evaluate a series of common segmentation methods. 
We evaluate top-10 and top-20 word segmentation, i.e. we take the top-10 and 20 words found in the segment boundaries of the train set to segment the test set. 
We also evaluate existing approaches like TextTiling \citep{hearst1997text}\footnote{We use the NLTK libary implementation of the algorithm \citep{bird2009natural}} and topic- and stage-segmentation methods from \citet{althoff2016large} and \citet{chen2020multi}, which segment discourse by topics and stages.
Lastly, we test zero-shot prompting long-context LLMs like GPT-4-turbo \citep{openai2024gpt} and the Claude variants Haiku, Sonnet, and Opus \citep{anthropic2024claude}.\footnote{These evaluations were performed in  May 2024.}
We omit open-source, instruct-tuned LLMs like Llama-2 \citep{touvron2023llama}, Llama-3 \citep{meta2024llama3}, or Mixtral \citep{jiang2024mixtral} because their context windows are not long enough for our transcripts.

We fit the topic and stage segmentation methods on our train split, and use three pre-trained encoders from Sentence-Transformers \citep{reimers-2019-sentence-bert}: the base-nli-stsb-mean-tokens (originally used in \citet{chen2020multi}), all-mpnet-base-v2, all-MiniLM-L12-v2.
These encoders did not vary in performance. Therefore, we report results on the first encoder and Appendix~\ref{sec:extended_results} reports the rest.
Stage segmentation requires the number of segments a priori; our experiments vary this to be either the rounded average or maximum number of segments in \datasetname{}.

\paragraph{Retrieval.}
We evaluate several methods for IR: 
Jaccard similarity \citep{jaccard1912distribution}, 
TF-IDF \citep{sammut2011encyclopedia}, 
BM25 \citep{robertson2009probabilistic}, 
ColBERTv2 \citep{santhanam2021colbertv2}, 
zero-shot prompting GPT-4-turbo, Claude Haiku, Claude Sonnet, and Claude Opus. 
Retrieval is challenging in our setting.
Retrieval methods must handle the semantic variability in how problems are discussed and referenced. 
The conversations do not follow a sequential order of problem IDs, and the references to problems are highly contextual, making lexical cues insufficient for dictionary-based retrieval.

A challenge in using traditional IR methods in our setting is specifying that nothing in the worksheet is linked to a segment, e.g., for informal or warm-up segments.
For instruct-tuned LLMs, we can simply specify this in the prompt.
For traditional IR methods, we must set a threshold value for what is deemed relevant enough to the segment. 
We perform 5-fold cross validation on the training set and set the threshold to the average value that best separates on the held-out fold. 
We report these thresholds in Appendix~\cref{sec:thresholds}.

\paragraph{POSR.}
We combine the best independent segmentation method with each retrieval method and report their joint performance.
We also evaluate zero-shot prompted GPT-4-turbo, Claude Haiku, Claude Sonnet, Claude Opus as POSR methods that perform segmentation and retrieval jointly.

\section{Results}
\label{sec:results}

\input{figures/posr_results}
\input{figures/segmentation_results}

\input{figures/segmentation_diff} 

Table~\ref{tab:posr} summarizes the joint evaluations, and Table~\ref{tab:segmentation_evaluations} summarizes the segmentation results.
\textbf{The POSR methods outperform most independent segmentation and retrieval approaches, and at lower costs.}
\texttt{POSR Opus} and \texttt{POSR GPT4} achieves slightly higher Line- and Time-SRS to their independent counterparts, and much higher to other combined independent approaches, e.g., \texttt{Opus}+\texttt{TFIDF} on both SRS metrics. 
Additionally, POSR methods are much more cost-effective, as they require only a single prompt to perform both segmentation and retrieval, rather than multiple prompts handling these tasks separately: \texttt{POSR Opus} and \texttt{POSR GPT4} cost \$11-\$21 per 100 transcripts, while the best combined independent methods, \texttt{Opus}+\texttt{GPT4}, cost \$54 per 100 transcripts.
This demonstrates the importance of jointly modelling segmentation and retrieval for better accuracy \textit{and} cost performance.
However, there is still room for improvement such as future work on developing and improving open-sourced long-context methods.

According to Table~\ref{tab:segmentation_evaluations}, \textbf{POSR methods perform better than most independent segmentation methods by a large margin.}
For example, \texttt{POSR Opus} improves upon topic and stage segmentation methods by $\sim57\%$ on $P_k$ and WindowDiff. 
The poor performance of top-10 and top-20 word segmentation indicates that segmentation cannot be solved by word-level cues alone. 
Additionally, we find that POSR methods perform better than their independent LLM segmentation counterparts.
For example, \texttt{POSR Sonnet} improves upon \texttt{Sonnet} across all segmentation metrics, such as $0.23 \rightarrow 0.13$ on Line-$P_k$ or $0.37 \rightarrow 0.31$ on Line-WindowDiff. 
Incorporating retrieval items enhances segmentation accuracy by providing additional context for more precise boundary detection, reinforcing the importance of treating segmentation and retrieval \textit{jointly}.

\textbf{The time- and line-based metrics for segmentation and SRS are well-correlated across methods}, indicating that accounting for time does not impact relative rankings. 
However, time-weighing is still important in  accounting for errors in long segments:
Time-$P_k$ errors are lower than Line-$P_k$ because it reduces the impact of oversegmentation whereas Time-WindowDiff amplifies errors from missing long segments.

\input{figures/logodds_use_case}

\begin{figure*}[t]
    \centering
    \includegraphics[width=0.9\linewidth]{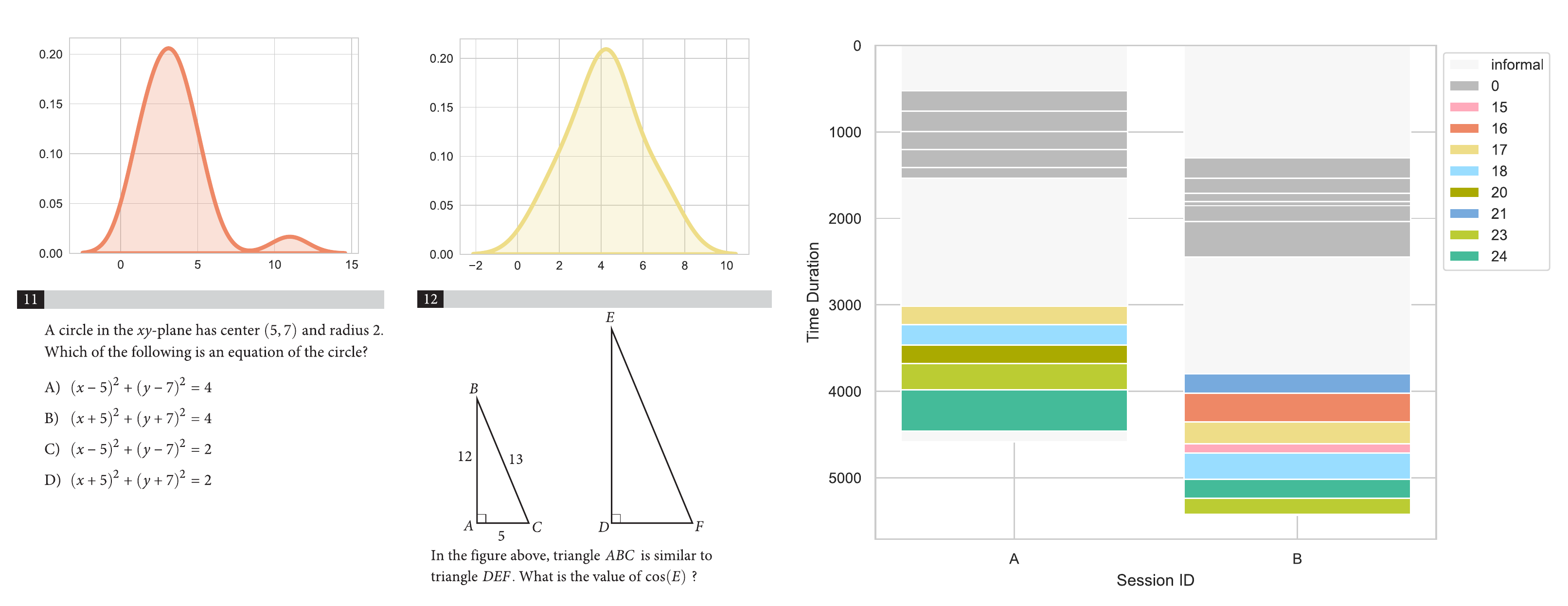}
    \caption{\small \textbf{Left:} Time spent (minutes) per worksheet problem. \textbf{Right:} Example of time management across two lessons. }
    \label{fig:example_489}
\end{figure*} 

\paragraph{Segmentation error analysis.}
To better understand sources of segmentation error, we investigate the difference in segment numbers (reported in Table~\ref{tab:segmentation_diff}) and we examine the bigram language in false segment insertions compared to true segment insertions with the log odds ratio, latent Dirichlet prior, measure defined in \citet{monroe2008fightin}. 
Table~\ref{tab:segmentation_diff} reveals that traditional methods oversegment, being sensitive to low-level topics shifts. 
Surprisingly, while \texttt{Haiku} has a higher segmentation error rate in Table~\ref{tab:posr}, it achieves the lowest segment count difference, altogether indicating that Haiku inserts new (albeit few) segments far away from true segment boundaries.
The log odds results in Table~\ref{tab:logodds_error} indicate that incorrect segments are inserted when the tutor introduces examples (e.g., \hlc[purple!30]{``let's say''}), alternative explanations (e.g., ``There are \hlc[violet!30]{different ways}  to solve this''), or participation prompts (e.g., ``how did you \hlc[teal!30]{like start} to approach this problem?'').
This analysis signals areas for improvement in precise segmentation.

\paragraph{Retrieval error analysis.}
We conduct a qualitative analysis on retrieval errors, particularly those in the independent  methods.
A large error source is caused by long segments that are incorrectly segmented for reasons illustrated in the previous section. 
For example, long problem segments are broken up and incorrectly linked.
Oversegmentation also yields shorter segment queries for retrieval, reducing the similarity to the target reference.
This particularly impacts traditional methods whose similarity thresholds are set with the ground truth segments as explained in Appendix~\ref{sec:thresholds}. 
In Appendix \ref{sec:extended_error_analysis}, we compare retrieval methods on \textit{ground-truth segments} and confirm that ground truth segments significantly boosts retrieval accuracy, especially for LLM methods.
Thus, we conclude that inaccurate segmentation is a critical bottleneck to mitigating downstream retrieval errors.

\section{Downstream Applications \label{sec:applications}}
There are several applications that POSR enables for gaining insights into tutoring practices at scale.  We illustrate two.
One application is a language analysis to compare how tutors talk about the same problem with the long vs. short talk times (top and bottom quartile). 
We use the log odds ratio measure from \citet{monroe2008fightin} to estimate the distinctiveness of a bigram using Edu-ConvoKit \citep{wang2024convokit}.
We report the top-3 bigrams on the most popular problem from \datasetname{} and qualitative examples in Figure~\ref{fig:qualitative_logodds_application}.
The log-odds analysis reveals that in short segments, tutors tend to stick to the language from the \hlc[orange!30]{``problem statement''} and immediately explain the answer. 
However, in longer segments, tutors provide  examples to students (e.g., \hlc[cyan!30]{``let's say''}), and offer conceptual explanations inferring the underlying mathematical concept (e.g., \hlc[cyan!30]{``this is a conditional probability question''}).
The second POSR application is the analysis of talk time distributions across different tutors and problems, such as in Figure~\ref{fig:example_489}: some problems have very different talk times (e.g., problem 11), while others have similar talk times (e.g., problem 12). 
Altogether, POSR enables these downstream applications and can tackle the large challenge of lesson structuring in education.

\section{Discussion and Conclusion}
We introduce the Problem-Oriented Segmentation and Retrieval (POSR), a task that jointly segments conversations and retrieves the problem discussed in each segment. 
We contribute the \datasetname{} dataset as a concrete case study of POSR in education.
\datasetname{} is the first large-scale dataset of tutoring conversations linked with worksheets, featuring 3,500 segments, 116 linked SAT\textregistered{} math problems and over 24,300 minutes of instruction.
To evaluate the joint performance and account for time in segmentation, we introduce the Segmentation and Retrieval Score (SRS) and time-based segmentation metrics for $P_k$ and WindowDiff. 
Our comprehensive evaluations highlight the importance of jointly modeling segmentation and retrieval, rather than treating them as independent tasks:  
POSR methods significantly outperform the independent approaches as measured against the traditional segmentation, SRS, and new time-based metrics. 
The LLM-based POSR methods achieve the best performance, but come at a higher cost, motivating future work on cost-effective solutions.
We also demonstrate the potential of POSR by showcasing downstream applications, such as a language analysis comparing tutoring strategies. 
In conclusion, our work establishes POSR as an important task to study conversation structure. 
The \datasetname{} dataset and the proposed methods pave the way for further research in joint segmentation and retrieval, with broad implications for educational technology, conversational analysis, and beyond.


\section{Limitations}
While our work provides a useful starting point for understanding conversations (such as in education) at scale, there are limitations to our work. 
Addressing these limitations will be an important area for future research. 

One limitation is the lack of connection to outcomes. 
While prior works have explored the relationship between duration and sequencing of problems on student attention (e.g., \citet{stevens2012role} \textit{inter alia}), there is limited research on how these factors impact long-term student learning, particularly in group-based settings.
Understanding this connection is crucial for grounding POSR in real contexts. 

Additionally, POSR does not rigorously link the language content  with the segment duration or ordering. 
This applies to other conversation domains as well, beyond education settings.
Linking content and quality of the language with the time allocation and sequencing matters \citep{suresh2018using}: Are tutors soliciting student contributions, or talking all the time? Are they restating or engaging with student contributions?  
While our downstream applications illustrate one form of language analysis with a log odds analysis, future work should investigate using language categories, instead of unsupervised methods for understanding language patterns.

Another limitation is the absence of audio and visual inputs. 
Our current models rely solely on textual data and miss non-verbal cues that add to the full context in understanding conversations. 
We also only use the problem text, and ignore the problem's visual components such as graph information.
Incorporating multimodal data, such as audio and visual inputs, could improve the accuracy of POSR systems. 

\section{Ethical Considerations}
The purpose of this work is to promote and improve effective interactions, such as in the setting of education, using NLP techniques.
The \datasetname{} dataset is intended for research purposes. 
The dataset should not be used for commercial purposes, and we ask that users of our dataset respect this restriction. 
As stewards of this data, we are committed to protecting the privacy and confidentiality of the individuals who contributed comments to the dataset. 
It is important to note that inferences drawn from the dataset should be interpreted with caution. 
The intended use case for this dataset is to further research on conversation interactions and education, towards the goal of improving interactions.
Unacceptable use cases include any attempts to identify users or use the data for commercial gain. 
We additionally recommend that researchers who do use our dataset take steps to mitigate any risks or harms to individuals that may arise.

\section*{Acknowledgments}
We are grateful to Schoolhouse.world for their contribution of data and domain expertise.
We are also grateful to the Schoolhouse.world team, Yann Hicke and Joy Yun for helpful discussions. 
\bibliography{custom}

\appendix

\section{$P_k$ and Time-$P_k$ \label{sec:pk}}

The $P_k$ metric is an established segmentation metric from \citet{beeferman1999statistical}. 
Similar to WindowDiff, it uses a line-based sliding window approach that measures boundary mismatches within the window. 
Lower values is better. 
For example, $P_k$ is computed as:
\begin{align*}
    P_k(Y, Y^*) &=  \\
    \frac{1}{N-k} &\sum^{N-k}_{j=1} \\
     \mathbbm{1}&\left(\mathbbm{1}{(b(s_{j:j+k})>0)} \neq  \mathbbm{1}{(b(s^*_{j:j+k}) > 0)} \right) 
\end{align*}
\vspace{-1em}

where $b(\cdot)$ represents the number of boundaries within the $\cdot$ window and $k$ is typically set to half of the average of the true segment line size.

Time-$P_k$ is calculated as:

\begin{align*}
    &\text{Time-}P_k(Y, Y^*) =  \\
    &\frac{1}{N-k} \sum^{N-k}_{j=1} \\
      \mathbbm{1}&\left(\mathbbm{1}{(b(s_{t_j^{\text{start}}:t_j^{\text{end}}+\Delta_k})>0)} \neq  \mathbbm{1}{(b(s^*_{t_j^{\text{start}}:t_j^{\text{end}}+\Delta_k}) > 0)} \right)
\end{align*}

where $\Delta_k$, the time duration of the sliding window, is half of the average true segment duration (similar to $k$).

\section{Prompts \label{sec:prompts}}

\input{prompts/segmentation_prompt}
\input{prompts/retrieval_prompt}
\input{prompts/segmentation_and_retrieval_prompt}

Recognizing that models are sensitive to prompt phrasing, we ran experiments on 15 transcripts to determine the best prompting approach for each task: independent segmentation, independent retrieval, and joint segmentation and retrieval. 
For each task, two authors collaboratively wrote a pool of prompt templates with varying phrasings. From these, we chose the top-performing template across all models to use for all transcripts.

\subsection{Independent segmentation}
For the independent segmentation task, we designed three distinct prompt templates:
\begin{enumerate}
    \item A template prompting the LLM to identify segments that each involve the discussion of an individual math problem, with an extra note emphasizing that each segment must involve the discussion of one math problem only;
    \item A template prompting the LLM to segment the transcript into contiguous segments, where each segment either involves (a) the discussion of a single math problem or (b) anything else (such as small talks, the introduction of the tutoring session, and the conclusion of the tutoring session, which, if contiguous, must be part of the same segment);
    \item A template prompting the LLM to detect lines where the tutor/students start transitioning to discussing a new math problem, as well as the line right after the tutor/students finish discussing the math problem, to mark the beginning of each segment
\end{enumerate}

We found that the first prompt template, shown in Figure~\ref{fig:segmentation_prompt}, performs best in terms of all segmentation metrics, i.e., WindowDiff and $P_k$ scores.

\subsection{Independent retrieval}

For the independent retrieval task, we designed two distinct prompt templates:

\begin{enumerate}
    \item A prompt template that retrieves for all segments in a transcript at once;
    \item A prompt template that retrieves for one segment at a time, independently for each segment.
\end{enumerate}

We found that both prompt templates perform comparably when given ground truth segments. 
However, when given imperfect, predicted segments, prompt template 2 performs significantly better in terms of SRS scores. 
We therefore choose to use prompt template 2, shown in  Figure~\ref{fig:retrieval_prompt}, for all transcripts.

\subsection{Joint segmentation and retrieval}

For the joint segmentation and retrieval task, we designed two distinct prompt templates:

\begin{enumerate}
    \item Similar to template 1 for the independent segmentation task, this template prompts the LLM to identify segments that each involve the discussion of an individual math problem, then determine which math problem was discussed in each segment or indicate if a math problem was discussed but not found in the provided set of problems.
    \item Similar to template 2 for the independent segmentation task, this template prompts the LLM to segment the transcript into contiguous segments, where each segment either involves (a) the discussion of a single math problem or (b) anything else (such as small talks, the introduction of the tutoring session, and the conclusion of the tutoring session, which, if contiguous, must be part of the same segment). It then requires determining if a math problem was discussed in each segment, and, if so, identifying the specific math problem or indicating if it can not be found in the provided set of problems.
\end{enumerate}

We found that the first prompt template, shown in Figure~\ref{fig:segmentation_and_retrieval_prompt}, performs best in terms of all relevant metrics, i.e., WindowDiff, $P_k$ scores, and SRS scores.

\section{Thresholds \label{sec:thresholds}}
A challenge in using traditional IR methods in our setting is specifying that nothing in the worksheet is linked to a segment, e.g., for informal or warm-up segments.
For traditional IR methods, we must set a threshold to determine which scores indicate that a worksheet problem is relevant enough to a segment.
We perform 5-fold cross-validation on the training set, testing threshold values from 0 to 1 in 0.01 intervals on ground truth segments, to determine the threshold that yields the highest retrieval accuracy on the held-out fold. We then average the best thresholds from each fold to obtain the final threshold for each method.

Note that for \texttt{BM-25} and \texttt{ColBERT}, which have unbounded relevance scores, we normalized the raw scores within the top 10 results for each query (as each worksheet has at least 10 problems to retrieve from). 
This normalization adjusts the scores relative to the top results, making them comparable across different queries and allowing us to set a threshold that would apply consistently across queries. 
Without this normalization, the scores would only be meaningful within the context of a single query and not comparable across different queries.

The threshold values for each traditional IR method are as follows:

\begin{itemize}
    \item \texttt{Jaccard}: $0.11$
    \item \texttt{tfidf}: $0.40$
    \item \texttt{BM-25}: $0.19$
    \item \texttt{ColBERT}: $0.14$
\end{itemize}

\section{Extended Results \label{sec:extended_results}}

Table~\ref{tab:extended_segmentation_results} shows the extended segmentation results where we used three pre-trained encoders from Sentence-Transformers \citep{reimers-2019-sentence-bert}: the base-nli-stsb-mean-tokens (originally used in \citet{chen2020multi}), all-mpnet-base-v2, all-MiniLM-L12-v2.
As the Table shows, the encoders did not vary much in segmentation performance.

\begin{spacing}{0.66}
\begin{table*}[t]
    \centering
    \resizebox{\linewidth}{!}{%
            \def\arraystretch{1.15}
      \begin{tabular}{c|cc|cc}
        \toprule
       \multicolumn{1}{c}{\bf Method} & \multicolumn{2}{c}{\bf $P_k$ ($\downarrow$)} & \multicolumn{2}{c}{\bf WindowDiff ($\downarrow$)}   \\
       \multicolumn{1}{c}{\bf } & \multicolumn{1}{c}{\bf Sentence} & \multicolumn{1}{c}{\bf Time} & \multicolumn{1}{c}{\textbf{Sentence}} &  \multicolumn{1}{c}{\textbf{Time}}  \\
        \midrule
           \texttt{Top-10}&  $0.58 \pm 0.04$  & $0.28 \pm 0.16$  & $1.0 \pm 0.01$  & $1.0 \pm 0.0$  \\ 
           \texttt{Top-20}   & $0.58 \pm 0.04$ &  $0.28 \pm 0.16$ &  $1.0 \pm 0.0$  & $1.0 \pm 0.0$  \\ 
           \texttt{TextTiling} &  $0.58 \pm 0.05$  &  $0.27 \pm 0.16$  &  $0.90 \pm 0.11$  &  $0.94 \pm 0.07$ \\ 
        \midrule
           \texttt{Topic, mpnet}    & $0.58 \pm 0.04$  & $0.27 \pm 0.16$  & $1.0 \pm 0.02$  &  $0.99 \pm 0.01$ \\ 
           \texttt{Topic, minilm}   & $0.58 \pm 0.04$ & $0.27 \pm 0.16$  & $1.0 \pm 0.02$  & $1.0 \pm 0.01$  \\ 
           \texttt{Topic, base}     & $0.58 \pm 0.04$ &  $0.27 \pm 0.16$ & $1.0 \pm 0.02$  & $1.0 \pm 0.01$  \\ 
           \texttt{Stage, mpnet, avg} &  $0.58 \pm 0.05$  &  $0.28 \pm 0.16$ & $0.99 \pm 0.03$  & $1.0 \pm 0.01$ \\ 
           \texttt{Stage, minilm, avg}  & $0.58 \pm 0.04$ & $0.28 \pm 0.16$ &  $1.0 \pm 0.02$ & $1.0 \pm 0.01$ \\ 
           \texttt{Stage, base, avg} &  $0.58 \pm 0.04$ &  $0.28 \pm 0.16$ & $1.0 \pm 0.0$  & $1.0 \pm 0.0$  \\ 
           \texttt{Stage, minilm, max} &  $0.58 \pm 0.04$  &  $0.28 \pm 0.16$ &  $1.0 \pm 0.00$ & $1.0 \pm 0.00$ \\ 
           \texttt{Stage, mpnet, max}  & $0.58 \pm 0.04$ &  $0.28 \pm 0.16$ & $1.0 \pm 0.01$  & $1.0 \pm 0.00$ \\ 
           \texttt{Stage, base, max}  &   $0.58 \pm 0.04$  & $0.28 \pm 0.16$  &  $1.0 \pm 0.0$ & $1.0 \pm 0.0$ \\ 
        \midrule
           \texttt{GPT4} &   $0.20 \pm 0.10$  &  $0.25 \pm 0.17$  & $0.33 \pm 0.09$  & $0.52 \pm 0.15$  \\ 
           \texttt{Haiku}  &  $0.29 \pm 0.14$ &   $0.30 \pm 0.17$ &  $0.39 \pm 0.14$ & $0.55 \pm 0.16$  \\ 
           \texttt{Sonnet} &   $0.24 \pm 0.14$ &    $0.23 \pm 0.18$ & $0.37 \pm 0.15$ & $0.53 \pm 0.17$ \\ 
           \texttt{Opus}  & $ 0.15 \pm 0.10$ &  \yellowcell $\bf  0.11 \pm 0.09 $  & $ 0.29 \pm 0.13$  &  \yellowcell $\bf 0.44 \pm 0.18$ \\ 
        \midrule
           \multicolumn{1}{c|}{\texttt{POSR GPT4}} & $0.16 \pm 0.09$  &  $0.17 \pm 0.16$ & $0.32 \pm 0.08$ & $0.52 \pm 0.16$   \\
           \multicolumn{1}{c|}{\texttt{POSR Haiku}} &  $0.24 \pm 0.10$ &  $0.23 \pm 0.14$ &  $0.35 \pm 0.11$ & $0.52 \pm 0.18$ \\
           \multicolumn{1}{c|}{\texttt{POSR Sonnet}} & \yellowcell $\bf 0.13 \pm 0.08$  &  \yellowcell $\bf 0.11 \pm 0.13$ & $0.31 \pm 0.10$ & $0.49 \pm 0.17$ \\
           \multicolumn{1}{c|}{\texttt{POSR Opus}} &  \yellowcell $\bf 0.13 \pm 0.08$  & $0.12 \pm 0.13$ & \yellowcell $\bf 0.28 \pm 0.10$ &  \yellowcell $\bf 0.44 \pm 0.17$  \\
        \bottomrule
      \end{tabular}
      }
      \caption{\small 
      \textbf{Extended segmentation evaluations ($\downarrow$ better).}
      \label{tab:extended_segmentation_results}
      }
\end{table*}
\end{spacing}

\input{figures/retrieval_results}

\section{Extended Error Analysis \label{sec:extended_error_analysis}}
To assess why independently performing retrieval on top of segmentation does not perform as well as the joint POSR methods (rf. Table~\ref{tab:posr}), we need to isolate and analyze the retrieval errors.
Therefore, we additionally evaluate the retrieval performance conditioned on the ground truth segments in Table~\ref{tab:independent_retrieval}. 
We find that the LLM-based solutions typically perform better than traditional IR methods, and for \texttt{GPT-4} and \texttt{Claude-Opus} near ceiling.
Interestingly, we find that \texttt{Haiku} performs similarly on retrieval as simpler methods such as using \texttt{Jaccard} similarity of \texttt{tfidf}. 
In our qualitative analysis, we find \texttt{Haiku}'s errors are due to retrieving incorrect worksheet problems on warm-up segments.
This is also the most common error type of other LLM-based retrievers. 

\end{document}

%% file: figures/dataset_statistics.tex
\begin{table}[t]
  \centering
  \small
  \begin{tabular}{cl|c}
    \toprule
    \multicolumn{1}{c}{\textbf{Transcripts}} & \multicolumn{1}{l}{Total Transcripts} & $300$ \\
    \multicolumn{1}{c}{} & \multicolumn{1}{l}{Total Speakers} & $1377$ \\
    \multicolumn{1}{c}{} & \multicolumn{1}{l}{Total Segments} & $3576$ \\
    \multicolumn{1}{c}{} & \multicolumn{1}{l}{Mean Speakers Per Transcript} & $6.37$ \\
    \multicolumn{1}{c}{} & \multicolumn{1}{l}{Mean Segments Per Transcript} & $11.92$ \\
    \multicolumn{1}{c}{} & \multicolumn{1}{l}{Mean Problems Per Transcript} & $7.43$ \\
    \multicolumn{1}{c}{} & \multicolumn{1}{l}{Mean Lines Per Transcript} & $495.51$ \\
    \multicolumn{1}{c}{} & \multicolumn{1}{l}{Mean Duration (mins)} & $81.62$ \\
    \midrule
    \multicolumn{1}{c}{\textbf{Worksheets}} & \multicolumn{1}{l}{Total Worksheets} & $7$ \\
    \multicolumn{1}{c}{} & \multicolumn{1}{l}{Total Problems} & $116$ \\
    \bottomrule
  \end{tabular}
  \caption{
  \small \textbf{\datasetname{} dataset statistics. }
  \label{tab:dataset_statistics}}
\end{table}

%% file: figures/posr_results.tex
\begin{spacing}{0.66}
\begin{table}[t]
    \centering
    \resizebox{\linewidth}{!}{%
            \def\arraystretch{1.15}
      \begin{tabular}{c|c|cc|c}
        \specialrule{.2em}{.2em}{.2em}
       \multicolumn{2}{c}{} & \multicolumn{2}{c}{\bf POSR Metrics} & \multicolumn{1}{c}{\bf } \\
       \multicolumn{1}{c}{\bf Segmentation} & \multicolumn{1}{c}{\bf Retrieval} & \multicolumn{2}{c}{\bf SRS ($\uparrow$)} & \multicolumn{1}{c}{\bf Cost ($\downarrow$)} \\
       \multicolumn{1}{c}{\bf Method} & \multicolumn{1}{c}{\bf Method} & \multicolumn{1}{c}{\bf Line} & \multicolumn{1}{c}{\bf Time} & \multicolumn{1}{c}{\bf } \\
        \specialrule{.2em}{.2em}{.2em}
           \texttt{Opus} & \texttt{Jaccard}     & $0.62 \pm 0.19$ & $0.63 \pm 0.19$ & $17.16 \pm 4.81$ \\ 
           \texttt{Opus} & \texttt{TFIDF}       & $0.63 \pm 0.22$ & $0.63 \pm 0.22$ & $17.16 \pm 4.81$ \\ 
           \texttt{Opus} & \texttt{BM25}        & $0.52 \pm 0.23$ & $0.53 \pm 0.23$ & $17.16 \pm 4.81$ \\ 
           \texttt{Opus} & \texttt{ColBERT}     & $0.50 \pm 0.23$ & $0.50 \pm 0.23$ & $17.16 \pm 4.81$ \\ 
           \midrule 
           \texttt{Opus} & \texttt{GPT4}        & $0.87 \pm 0.13$ & $0.88 \pm 0.13$ & $54.32 \pm 14.71$ \\ 
           \texttt{Opus} & \texttt{Haiku}       & $0.58 \pm 0.23$ & $0.59 \pm 0.24$ & $18.17 \pm 4.90$ \\ 
           \texttt{Opus} & \texttt{Sonnet}      & $0.70 \pm 0.20$ & $0.71 \pm 0.20$ & $29.25 \pm 7.04$ \\ 
           \texttt{Opus} & \texttt{Opus}        & $0.85 \pm 0.11$ & $0.86 \pm 0.11$ & $77.58 \pm 22.54$ \\
        \specialrule{.2em}{.2em}{.2em}
           \multicolumn{2}{c|}{\texttt{POSR GPT4}}      & \yellowcell $\bf  0.88 \pm 0.12$ & \yellowcell $ \bf 0.89 \pm 0.11$ & $11.72 \pm 2.71$ \\
           \multicolumn{2}{c|}{\texttt{POSR Haiku}}     & $0.61 \pm 0.22$ & $0.62 \pm 0.23$ &  \yellowcell $ \bf 0.35 \pm 0.08$ \\
           \multicolumn{2}{c|}{\texttt{POSR Sonnet}}    & $0.85 \pm 0.14$ & $0.85 \pm 0.14$ & $4.22 \pm 0.93$ \\
           \multicolumn{2}{c|}{\texttt{POSR Opus}}      & \yellowcell $\bf 0.88 \pm 0.11$ &  \yellowcell $\bf 0.89 \pm 0.11$ & $21.11 \pm 4.64$ \\
        \specialrule{.2em}{.2em}{.2em}
      \end{tabular}
      }
      \caption{\small 
      \textbf{POSR evaluations.} The best average is \hl{\textbf{highlighted}}.
      \label{tab:posr}
      }
\end{table}
\end{spacing}

%% file: figures/segmentation_results.tex
\begin{spacing}{0.66}
\begin{table}[t]
        \centering
    \resizebox{\linewidth}{!}{%
            \def\arraystretch{1.15}
      \begin{tabular}{c|cc|cc}
        \specialrule{.2em}{.2em}{.2em}
       \multicolumn{1}{c}{} & \multicolumn{4}{c}{\bf Segmentation Metrics} \\
       \multicolumn{1}{c}{\bf } &  \multicolumn{2}{c}{\bf $P_k$ ($\downarrow$)} & \multicolumn{2}{c}{\bf WindowDiff ($\downarrow$)} \\
       \multicolumn{1}{c}{\bf Method} & \multicolumn{1}{c}{\bf Line} & \multicolumn{1}{c}{\bf Time} & \multicolumn{1}{c}{\textbf{Line}} &  \multicolumn{1}{c}{\textbf{Time}}  \\
        \midrule
           \texttt{Top-10}&  $0.58 \pm 0.04$  & $0.28 \pm 0.16$  & $1.0 \pm 0.01$  & $1.0 \pm 0.0$  \\ 
           \texttt{Top-20}   & $0.58 \pm 0.04$ &  $0.28 \pm 0.16$ &  $1.0 \pm 0.0$  & $1.0 \pm 0.0$  \\ 
           \texttt{TextTiling} &  $0.58 \pm 0.05$  &  $0.27 \pm 0.16$  &  $0.90 \pm 0.11$  &  $0.94 \pm 0.07$ \\ 
           \texttt{Topic} &   $0.58 \pm 0.04$ &  $0.27 \pm 0.16$ & $1.0 \pm 0.02$  & $1.0 \pm 0.01$  \\ 
           \texttt{Stage}$_{\text{avg}}$  &  $0.58 \pm 0.04$ &  $0.28 \pm 0.16$ & $1.0 \pm 0.0$  & $1.0 \pm 0.0$  \\ 
           \texttt{Stage}$_{\text{max}}$  &   $0.58 \pm 0.04$  & $0.28 \pm 0.16$  &  $1.0 \pm 0.0$ & $1.0 \pm 0.0$ \\ 
        \midrule
           \texttt{GPT4} &   $0.20 \pm 0.10$  &  $0.25 \pm 0.17$  & $0.33 \pm 0.09$  & $0.52 \pm 0.15$  \\ 
           \texttt{Haiku}  &  $0.28 \pm 0.13$ &   $0.30 \pm 0.17$ &  $0.38 \pm 0.13$ & $0.54 \pm 0.16$  \\ 
           \texttt{Sonnet} &   $0.23 \pm 0.14$ &    $0.22 \pm 0.17$ & $0.37 \pm 0.14$ & $0.54 \pm 0.17$ \\ 
           \texttt{Opus}  & $ 0.15 \pm 0.10$ &  \yellowcell $\bf  0.11 \pm 0.09 $  & $ 0.29 \pm 0.13$  &  \yellowcell $\bf 0.44 \pm 0.18$\\ 
        \specialrule{.2em}{.2em}{.2em}
           \multicolumn{1}{c|}{\texttt{POSR GPT4}} & $0.16 \pm 0.09$  &  $0.17 \pm 0.16$ & $0.32 \pm 0.08$ & $0.52 \pm 0.16$   \\
           \multicolumn{1}{c|}{\texttt{POSR Haiku}} &  $0.24 \pm 0.10$ &  $0.23 \pm 0.14$ &  $0.35 \pm 0.11$ & $0.52 \pm 0.18$ \\
           \multicolumn{1}{c|}{\texttt{POSR Sonnet}} & \yellowcell $\bf 0.13 \pm 0.08$  &  \yellowcell $\bf 0.11 \pm 0.13$ & $0.31 \pm 0.10$ & $0.49 \pm 0.17$ \\
           \multicolumn{1}{c|}{\texttt{POSR Opus}} &  \yellowcell $\bf 0.13 \pm 0.08$  & $0.12 \pm 0.13$ & \yellowcell $\bf 0.28 \pm 0.10$ &  \yellowcell $\bf 0.44 \pm 0.17$  \\
        \specialrule{.2em}{.2em}{.2em}
      \end{tabular}
      }
      \caption{\small 
      \textbf{Segmentation evaluations. } The best average is \hl{\textbf{highlighted}}.
      \label{tab:segmentation_evaluations}
      }
\end{table}
\end{spacing}

%% file: figures/segmentation_diff.tex
\definecolor{lightlavender}{RGB}{230,230,250}

\begin{table*}[ht]
    \centering
    \begin{minipage}[t]{0.21\textwidth}
        \centering
        \resizebox{\linewidth}{!}{%
                \def\arraystretch{1.15}
          \begin{tabular}{c|c}
            \toprule
           \multicolumn{1}{c}{\bf Method} & \multicolumn{1}{c}{\bf \# Segment Diff}  \\
            \midrule
               \texttt{Top-10} & $236.84 \pm 75.98$ \\ 
               \texttt{Top-20} & $305.37 \pm 90.04$ \\ 
               \texttt{TextTiling} & $42.97 \pm 17.93$\\ 
               \texttt{Topic} & $148.61 \pm 52.044$ \\ 
               \texttt{Stage$_\text{avg}$} &  $367.40 \pm 115.82$\\ 
               \texttt{Stage$_\text{max}$} & $371.27 \pm 118.84$ \\ 
            \midrule
               \texttt{GPT4} &  $-1.24 \pm 4.51$ \\ 
               \texttt{Haiku} &  $0.73 \pm 4.90$  \\ 
               \texttt{Sonnet}  &  $2.86 \pm 5.50$  \\ 
               \texttt{Opus} & $4.82 \pm 5.86$  \\ 
            \midrule
               \texttt{POSR GPT4} &  $1.09 \pm 4.47$ \\ 
               \texttt{POSR Haiku} &  $1.02 \pm 4.02$  \\ 
               \texttt{POSR Sonnet}  & $3.67 \pm 3.8$ \\ 
               \texttt{POSR Opus} & $2.64 \pm 3.64$ \\ 
            \bottomrule
          \end{tabular}
          }
          \caption{\small \label{tab:segmentation_diff} \textbf{Difference in number of segments.} }
    \end{minipage}%
    \hfill
    \begin{minipage}[t]{0.70\textwidth}
          \centering
          \resizebox{\linewidth}{!}{%
            \def\arraystretch{1.00}
          \begin{tabular}{lll}
            \toprule
            \textbf{Category} & \textbf{Bigram (log odds)}  \\ 
            \midrule
            \bf Providing Examples & \hlc[purple!30]{``lets\_say''} (2.26), \hlc[purple!30]{``yeah\_say''} (1.51) \\
            \multicolumn{2}{l}{e.g., \hlc[purple!30]{Let’s say} we have the function X squared plus 5 x plus 6.}  \\
            \midrule
            \bf Alternative explanations & \hlc[violet!30]{``differ\_way''} (1.50), \hlc[violet!30]{``simpler\_way''} (1.23) \\
            \multicolumn{2}{l}{e.g., There are \hlc[violet!30]{different ways} to solve this as well.}  \\
            \midrule
            \bf Prompting participation & \hlc[teal!30]{like\_start} (1.51), \hlc[teal!30]{try\_find} (1.48),  \hlc[teal!30]{guy\_know} (1.48) \\
            \multicolumn{2}{l}{e.g., So, [STUDENT], how did you \hlc[teal!30]{like start} to approach this problem?}  \\
            \bottomrule
          \end{tabular}
          }
          \caption{\small \textbf{Bigram categories founded in falsely inserted boundaries by \texttt{POSR Opus}}. Incorrect segments are inserted when the tutor provides examples (\hlc[purple!30]{``let's\_say''}), alternative explanations (\hlc[violet!30]{``diff\_way''}), or prompts for participation (\hlc[teal!30]{``like\_start''}). \label{tab:logodds_error}}
    \end{minipage}
\end{table*}

%% file: figures/logodds_use_case.tex
\begin{figure*}[t]
    \centering
    \begin{minipage}[t]{0.30\textwidth}
        \centering
        \includegraphics[width=\linewidth]{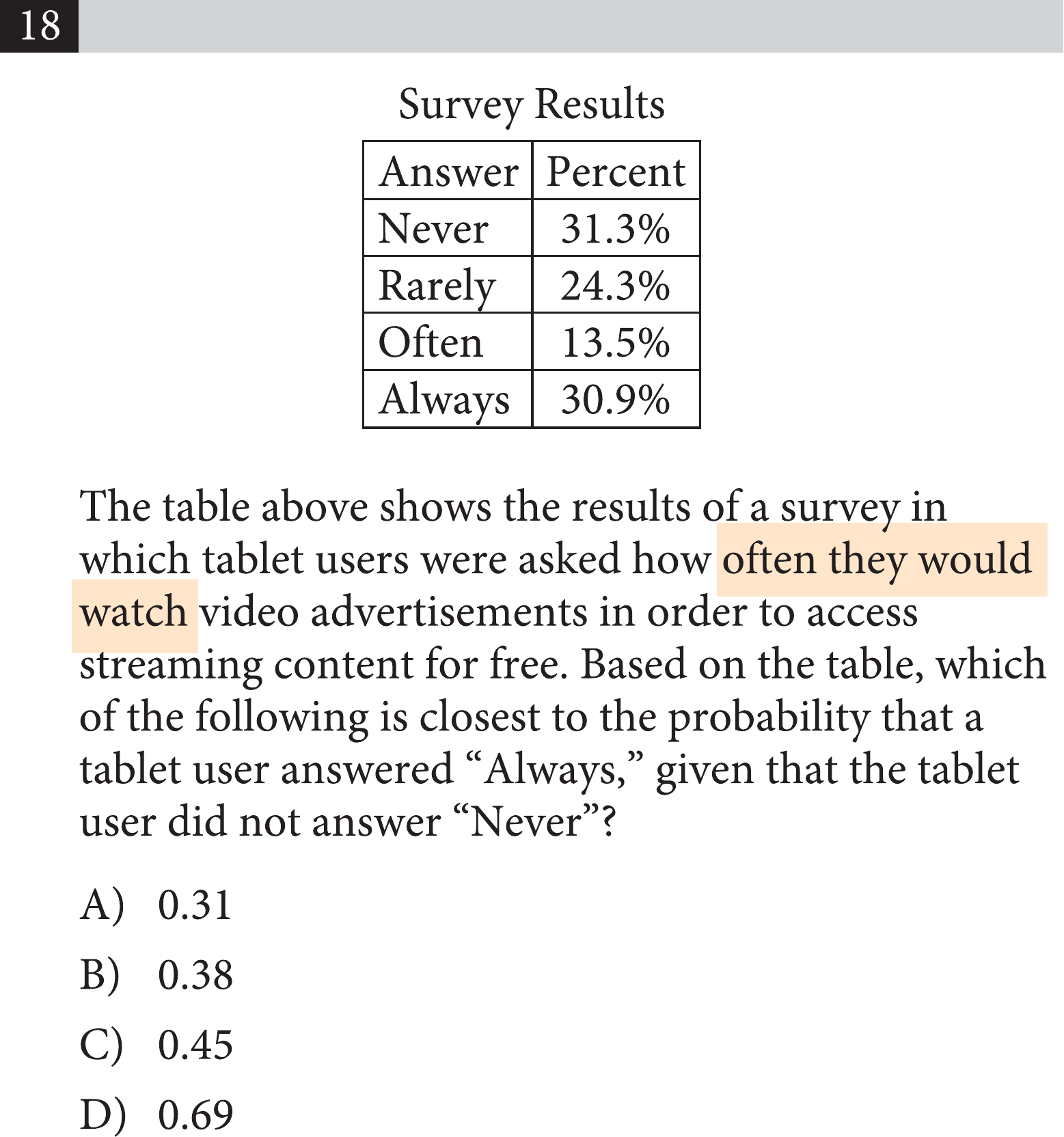}
    \end{minipage}%
    \hfill
    \begin{minipage}[t]{0.70\textwidth}
          \centering
            \vspace{-13em}
          \resizebox{0.90\textwidth}{!}{%
            \def\arraystretch{1.00}
          \begin{tabular}{l|l}
                \specialrule{.2em}{.2em}{.2em}
                \textbf{Long segments} & \hlc[cyan!30]{let\_see} ($0.683$), \hlc[cyan!30]{let\_say} ($0.683$), \hlc[cyan!30]{conditional\_probability} ($0.602$) \\
                \midrule
                \textbf{Example} & Tutor: And then someone wants to take a look at Question 18 [...] \\
                                   & you might deal with something called \hlc[cyan!30]{conditional probability.} Right?  \\
                                   & So \hlc[cyan!30]{conditional probability} means what is the probability of something  \\ 
                                   & occurring when something else doesn't occur. So \hlc[cyan!30]{let's say} that you have \\ 
                                   & 2 events A and B. The probability that a occurs assuming that B occurs \\
                                   & which we denote like this probability of A assuming B [...] so \hlc[cyan!30]{let's say} \\
                                   & that we have some event a. and we have some event. B. So a. And then we [..] \\
                \specialrule{.2em}{.2em}{.2em}
                \textbf{Short segments} & \hlc[orange!30]{always\_divided} ($2.025$), \hlc[orange!30]{often\_would} ($1.658$), \hlc[orange!30]{would\_watch} ($1.658$)\\
                \midrule
                \textbf{Example} & Tutor: So now 18. \hlc[orange!30]{[..reading aloud the problem..]} So let's just take \\ 
                                   & 31.3. Take that off of a 100, so 68, point 7. That's going to be 30.  \\ 
                                   & Point 9, over 68.7, which i'm guessing is around point 4, 5, just to  \\
                                   & guess. based off of the answer choices. Yep. The answer is, See that's \\
                                   & pretty much all there is to that problem. You just have to get rid of this. \\
                \specialrule{.2em}{.2em}{.2em}
          \end{tabular}
          }
    \end{minipage}
    \caption{\small 
    \textbf{Qualitative examples \& log odds.} We report the top-3 bigrams in segments talking about the left problem. We compare long segments (top quartile duration) and short segments (bottom quartile duration). 
    Longer segments tend to provide conceptual explanations (\hlc[cyan!30]{``let's say''}, \hlc[cyan!30]{conditional probability}).
    Shorter segments tend to stick more to the \hlc[orange!30]{problem} at hand.
    \label{fig:qualitative_logodds_application}}
\end{figure*}

%% file: prompts/segmentation_prompt.tex
\begin{figure*}[h]
    \centering
    \small
    \begin{tcolorbox}[
    prompt,
    title={\textbf{Independent Segmentation Prompt}},
    ]
    \#\#\# System:\\
You are an assistant who will be given a transcript of an SAT math tutoring session between a tutor and a group of students. Each line in the transcript will contain the line index, the speaker (tutor or student), and the utterance. Your job is to read the transcript and identify segments that each involve the discussion of an individual math problem. Note that each segment must involve the discussion of one math problem only. 
\\ \\
Please then output the first line index and last line index of each segment as a list of lists: \\
\text{[[<first line index of segment 1>, <last line index of segment 1>], ...,} \\
\text{[<first line index of segment n>, <last line index of segment n>]]}.
\\ \\
Only output a list of lists. Do not output any additional text or explanations.\\ \\
\#\#\# User: \\
Please read the transcript below and identify segments that each involve the discussion of an individual math problem: \\
\{transcript\}
\\ \\
Please output the first line index and last line index of each segment as a list of lists: \text{[[<first line index of segment 1>, <last line index of segment 1>], ...,} \\
\text{[<first line index of segment n>, <last line index of segment n>]]}. 
\\ \\
Only output a list of lists. Do not output any additional text or explanations.

    \end{tcolorbox}
    \caption{
    \textbf{Prompt for the independent segmentation task for LLM methods.}
    \texttt{\{transcript\}} is the placeholder for the entire tutoring transcript whose lines have the following format: \texttt{\{idx\} \{speaker\}: \{utterance\}}.
    \label{fig:segmentation_prompt}}
\end{figure*}

%% file: prompts/retrieval_prompt.tex
\begin{figure*}[h]
    \centering
    \small
    \begin{tcolorbox}[
    prompt,
    title={\textbf{Independent Retrieval Prompt}},
    ]
\#\#\# System:\\
You are an assistant who will be given (1) a segment of an SAT math tutoring session between a tutor and a group of students and (2) the set of math problems that might be discussed in the segment. Your job is to read the segment's transcript and set of math problems, then determine the math problem that was discussed in the segment, if any. If no math problem was discussed in the segment, please output "null". If a math problem was discussed in the segment but not found in the provided set of problems, please output -1. If a math problem was discussed in the segment and is found in the provided set of problems, please output the ID of the problem. Please do not output any additional text or explanations.
\\ \\ 

\#\#\# User: \\
Please read the segment’s transcript, read the set of math problems that might be discussed in the segment, and determine the math problem that was discussed in the segment, if any.
\\ \\ 
Segment: \\
\{transcript\}
\\ \\ 
Math problems: \\
\{problems\}
\\ \\
If no math problem was discussed in the segment, please output "null". If a math problem was discussed in the segment but not found in the provided set of problems, please output -1. If a math problem was discussed in the segment and is found in the provided set of problems, please output the ID of the problem. Please do not output any additional text or explanations.
    \end{tcolorbox}
    \caption{
    \textbf{Prompt for the independent retrieval task for LLM methods.}
    \texttt{\{transcript\}} is the placeholder for a tutoring segment's transcript whose lines have the following format: \texttt{\{speaker\}: utterance}. \texttt{\{problems\}} is the placeholder for the worksheet problems relevant to the session that have the following format: \texttt{Problem ID \{id\}: problem string}.
    \label{fig:retrieval_prompt}}
\end{figure*}

%% file: prompts/segmentation_and_retrieval_prompt.tex
\begin{figure*}[h]
    \centering
    \small
    \begin{tcolorbox}[
    prompt,
    title={\textbf{Segmentation and Retrieval Prompt}},
    ]
\#\#\# System:\\
You are an assistant who will be given (1) a transcript of an SAT math tutoring session between a tutor and a group of students and (2) the set of math problems that might be discussed in the session. Each line in the transcript contains the line index, the speaker (tutor or student), and the utterance. Each math problem corresponds to a problem ID.
\\ \\
Your first job is to read the transcript and identify segments that each involve the discussion of an individual math problem. Note that each segment must involve the discussion of one math problem only. Your second job is to determine the math problem that was discussed in each of the segments you identified. Please then output the first line index and last line index of each segment, along with the ID of the problem discussed in each segment as a list of JSON objects: \\
\texttt{[\{"start\_line\_idx": <first line index of segment 1>, "end\_line\_idx": <last line index of segment 1>, "problem\_id": <ID of problem discussed in segment 1>\}, ..., \{"start\_line\_idx": <first line index of segment n>, "end\_line\_idx": <last line index of segment n>, "problem\_id": <ID of problem discussed in segment n>\}]}.
\\ \\
If a math problem was discussed in a segment but not found in the provided set of problems, let the problem\_id be -1. Only output the list of JSON objects. Do not output any additional text or explanations.
\\ \\
\#\#\# User: \\
Please read the transcript, identify segments that each involve the discussion of an individual math problem, and determine the math problem that was discussed in each of the segments you identified.
\\ \\
Transcript: \\
\{transcript\}
\\ \\
Math problems: \\
\{problems\}
\\ \\
Please output the first line index and last line index of each segment, along with the ID of the problem discussed in each segment as a list of JSON objects: \\ \texttt{[\{"start\_line\_idx": <first line index of segment 1>, "end\_line\_idx":
<last line index of segment 1>, "problem\_id": <ID of problem discussed in segment 1>\}, ..., \{"start\_line\_idx": <first line index of segment n>, "end\_line\_idx": <last line index of segment n>, "problem\_id": <ID of problem discussed in segment n>\}]}.
\\ \\
If a math problem was discussed in a segment but not found in the provided set of problems, let the problem\_id be -1. Only output the list of JSON objects. Do not output any additional text or explanations.
    \end{tcolorbox}
    \caption{
    \textbf{Prompt for the joint segmentation and retrieval task for LLM methods.}
    \texttt{\{transcript\}} is the placeholder for the entire tutoring transcript whose lines have the following format: \texttt{\{idx\} \{speaker\}: \{utterance\}}. \texttt{\{problems\}} is the placeholder for the worksheet problems relevant to the session that have the following format: \texttt{Problem ID \{id\}: problem string}.
    \label{fig:segmentation_and_retrieval_prompt}}
\end{figure*}

%% file: figures/retrieval_results.tex
\begin{spacing}{0.66}
\begin{table}[t]
    \centering
    \resizebox{\linewidth}{!}{%
            \def\arraystretch{1.15}
      \begin{tabular}{c|c}
        \toprule
       \multicolumn{1}{c}{\bf Method} & \multicolumn{1}{c}{\bf Accuracy $\uparrow$}   \\
        \midrule 
           \texttt{Jaccard} & $0.64 \pm 0.20$  \\
          \texttt{tfidf} & $0.68 \pm 0.20$  \\
           \texttt{BM-25} & $0.51 \pm 0.22$  \\
           \texttt{ColBERT} & $0.51 \pm 0.22$ \\
        \midrule
           \texttt{GPT-4} & \yellowcell $\bf 0.97 \pm 0.07$ \\
           \texttt{Claude Haiku} & $0.69 \pm 0.26$ \\
           \texttt{Claude Sonnet}  & $0.86 \pm 0.16$ \\
           \texttt{Claude Opus} & $0.95 \pm 0.09$  \\
        \bottomrule
      \end{tabular}
      }
      \caption{\small 
      \textbf{Independent retrieval evaluations on the ground truth segments.}
      \label{tab:independent_retrieval}
      }
\end{table}
\end{spacing}